\documentclass[runningheads]{llncs}
\usepackage[T1]{fontenc}
\usepackage{pifont}
\usepackage{amssymb, amsmath, amsfonts}
\usepackage{euscript}
\usepackage{mathtools}
\usepackage{subcaption}
\usepackage{graphicx}
\usepackage{caption}
\usepackage{subcaption}  
\usepackage{booktabs}
\usepackage{microtype}
\usepackage{comment}
\usepackage{graphicx}
\usepackage[table]{xcolor}
\usepackage{hyperref}
\graphicspath{Figures/}
\usepackage{multirow}
\usepackage{adjustbox}
\hypersetup{
    colorlinks=true,
    linkcolor=blue,
    filecolor=cyan,      
    urlcolor=magenta,
    }
\usepackage{graphicx,verbatim}
\usepackage{amsmath}
\usepackage{pifont}
\usepackage{booktabs}
\usepackage{multirow} 
\usepackage{array}
\usepackage{tabularx} 
\usepackage{xcolor}
\usepackage{graphicx}
\usepackage{makecell}
\usepackage{mathtools} 
\usepackage[capitalise,noabbrev]{cleveref}
\usepackage{adjustbox}

\newcolumntype{L}[1]{>{\raggedright\arraybackslash}p{#1}}
\usepackage{color}

\begin{document}
\title{WaRA: Wavelet Low-Rank Adaptation for Medical Image Classification}
%

\author{
Moein Heidari\inst{1} \and
Yijin Huang\inst{1,2} \and
Yasamin Medghalchi\inst{1} \and
Alireza Rafiei\inst{1} \and
Roger Tam\inst{1} \and
Ilker Hacihaliloglu\inst{1}
}
\authorrunning{Heidari et al.}
\institute{
The University of British Columbia, Vancouver, BC, Canada \and
Southern University of Science and Technology, Shenzhen, China \\
\email{moein.heidari@ubc.ca}
}

\maketitle              
\begin{abstract}
Adapting large pretrained vision models to medical image classification is often limited by memory, computation, and task-specific specializations. Parameter-efficient fine-tuning (PEFT) methods like LoRA reduce this cost by learning low-rank updates, but operating directly in feature space can struggle to capture the localized, multi-scale features common in medical imaging. We propose WaRA, a wavelet-structured adaptation module that performs low-rank adaptation in a wavelet domain. WaRA reshapes patch tokens into a spatial grid, applies a fixed discrete wavelet transform, updates subband coefficients using a shared low-rank adapter, and reconstructs the additive update through an inverse wavelet transform. This design provides a compact trainable interface while biasing the update toward both coarse structure and fine detail. For extremely low-resource settings, we introduce Tiny-WaRA, which further reduces trainable parameters by learning only a small set of coefficients in a fixed basis derived from the pretrained weights through a truncated SVD. Experiments on medical image classification across four modalities and datasets demonstrate that WaRA consistently improves performance over strong PEFT baselines, while retaining a favorable efficiency profile. Our code is publicly available at~\href{https://github.com/moeinheidari7829/WaRA}{\textcolor{magenta}{GitHub}}.


\keywords{Parameter-efficient fine-tuning \and Low-rank adaptation \and Pre-trained models.}

\end{abstract}
\section{Introduction}
\noindent
Medical image classification has advanced substantially with deep learning, yet
modern foundation models are data-intensive, while curating large annotated
medical datasets remains costly and constrained by privacy and class imbalance.
Transfer learning and Parameter-Efficient Fine-Tuning (PEFT) have therefore become
central strategies for adapting foundation models to specialized medical tasks
while mitigating overfitting \cite{li2022cross,fu2023effectiveness}. PEFT methods typically freeze the pre-trained model and adapt only a small set of parameters, making them highly relevant for medical domains with limited data and specialized institutional settings \cite{chen2025personalized}. Representative directions include selective fine tuning \cite{guo2021parameter}, adapter based methods \cite{houlsby2019parameter,pfeiffer2020adapterfusion}, and prompt based tuning \cite{lester2021power,jia2022visual}.

\noindent A prominent PEFT approach is low-rank adaptation (LoRA)~\cite{hu2022lora}, which augments a frozen weight matrix $W_0$ with a low-rank update $\Delta W = BA$.
This mechanism has been extended to control parameter growth in medical imaging
settings, including continual segmentation and personalized federated
learning \cite{chen2024low,medghalchi2025synthetic,chen2025personalized}. In this work, we focus on medical image classification, where adapting large visual backbones efficiently is important for small datasets and decentralized clinical deployment.  Despite its empirical success, standard LoRA has notable limitations. First, most low-rank adapters operate directly in feature space \cite{lialin2023scaling}. Moreover, in extreme resource-limited regimes—common in decentralized healthcare deployments—even conventional LoRA-style adapters can be non-trivial to store, distribute, and load, incurring storage and bandwidth overhead \cite{gao2024parameter}.
\\
Recent works explore reparameterizing adaptation in a transformed domain to induce more structured updates. In the frequency domain, FouRA learns low-rank adaptations via Fourier projections \cite{borse2024foura}, FourierFT explores compressing trainable updates via discrete Fourier parameterizations \cite{gao2024parameter}, and LoCA proposes selective frequency component learning \cite{duloca}. These works suggest that frequency parameterizations can provide structured update spaces. For medical image classification, such structure is important because subtle variations and high visual similarity across categories create both intra-class variation and inter-class similarity \cite{huo2024hifuse}. Effective adaptation should therefore integrate multi-scale information, capturing fine local details while retaining global context.
This motivates wavelet-based alternatives, which, unlike the global Fourier basis, are localized in both space and frequency. Wavelets have been integrated into
scattering networks \cite{bruna2013invariant}, transformer architectures
\cite{yao2022wave}, and generative modeling
\cite{mattar2024wavelets}, demonstrating that multi-scale
decompositions can improve both efficiency and fidelity.
\\
In this work, we propose WaRA (Wavelet Low-Rank Adaptation), a PEFT approach designed for medical image classification that performs adaptation in a wavelet domain to exploit multi-scale structure. WaRA decomposes intermediate representations into frequency subbands that capture complementary information, performs low-rank updates in this transformed space, and reconstructs in the original domain. This wavelet-structured parameterization biases adaptation toward both coarse structure (low-frequency) and fine-grained features (high-frequency) that are prominent in medical imaging. We further develop Tiny-WaRA for extremely low-resource settings, which reduces the trainable budget by parameterizing the adapter update with a compact trainable core while keeping a fixed basis derived from pre-trained weights \cite{balazy2024lora,morris2026learning}.
\\
In summary, our key contributions are: (1) We present WaRA, a wavelet-domain low-rank adaptation module that leverages multi-scale subband decomposition to capture both global structure and fine-grained detail for medical image classification. (2) We develop Tiny-WaRA for extremely low-resource settings, substantially reducing trainable parameters compared to baseline frequency-domain methods by learning a compact core update in a fixed basis derived from pre-trained weights. (3) We conduct extensive experiments on medical image classification benchmarks, complemented by targeted ablations and frequency-domain analyses that elucidate when and why wavelet-structured adaptation is effective.

\section{Method}

\begin{figure}[t]
\centering
\begin{subfigure}[b]{0.43\linewidth}
  \centering
  \includegraphics[width=\linewidth]{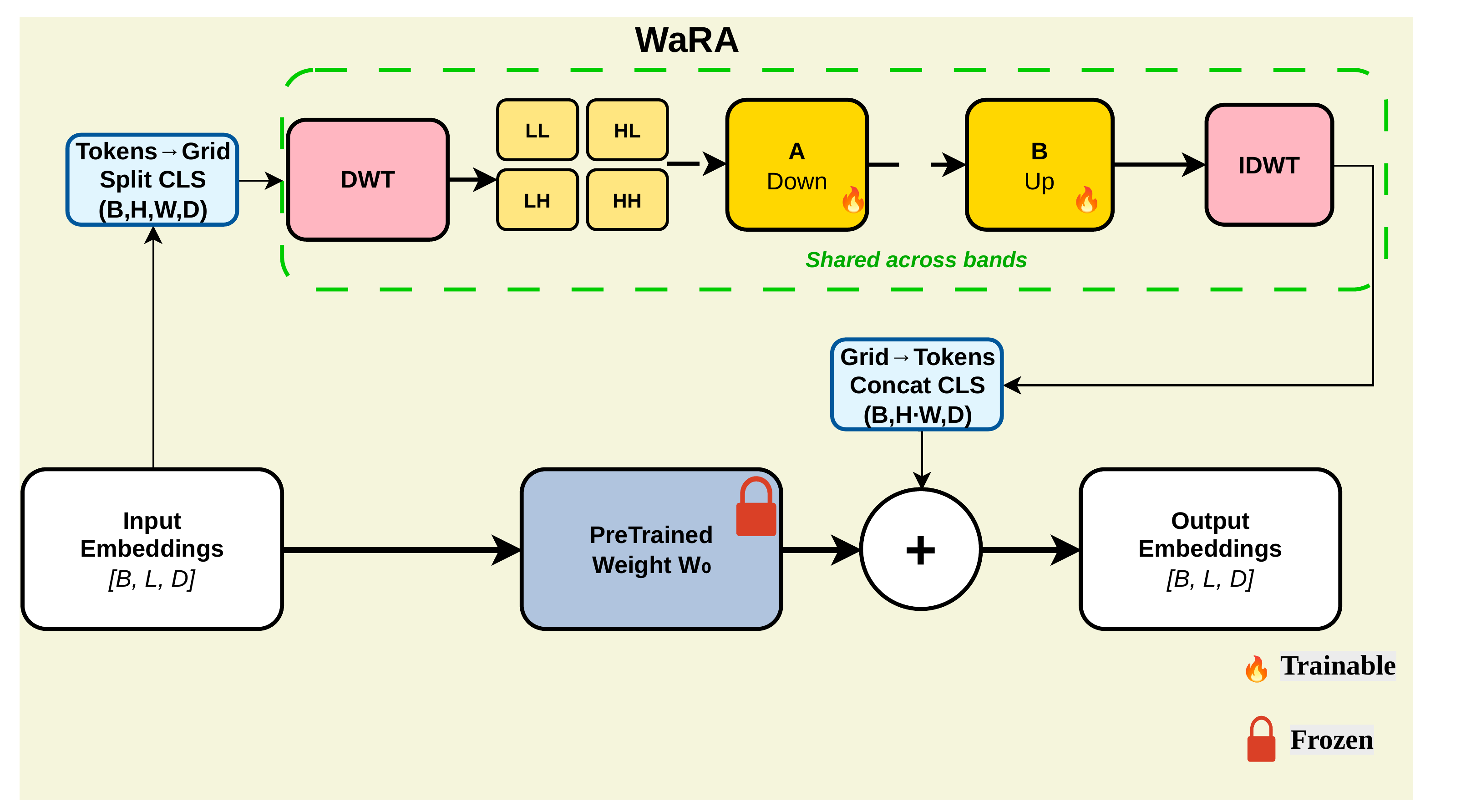}
  \caption{WaRA.}
  \label{fig:overview_wara}
\end{subfigure}
\hfill
\begin{subfigure}[b]{0.54\linewidth}
  \centering
  \includegraphics[width=\linewidth]{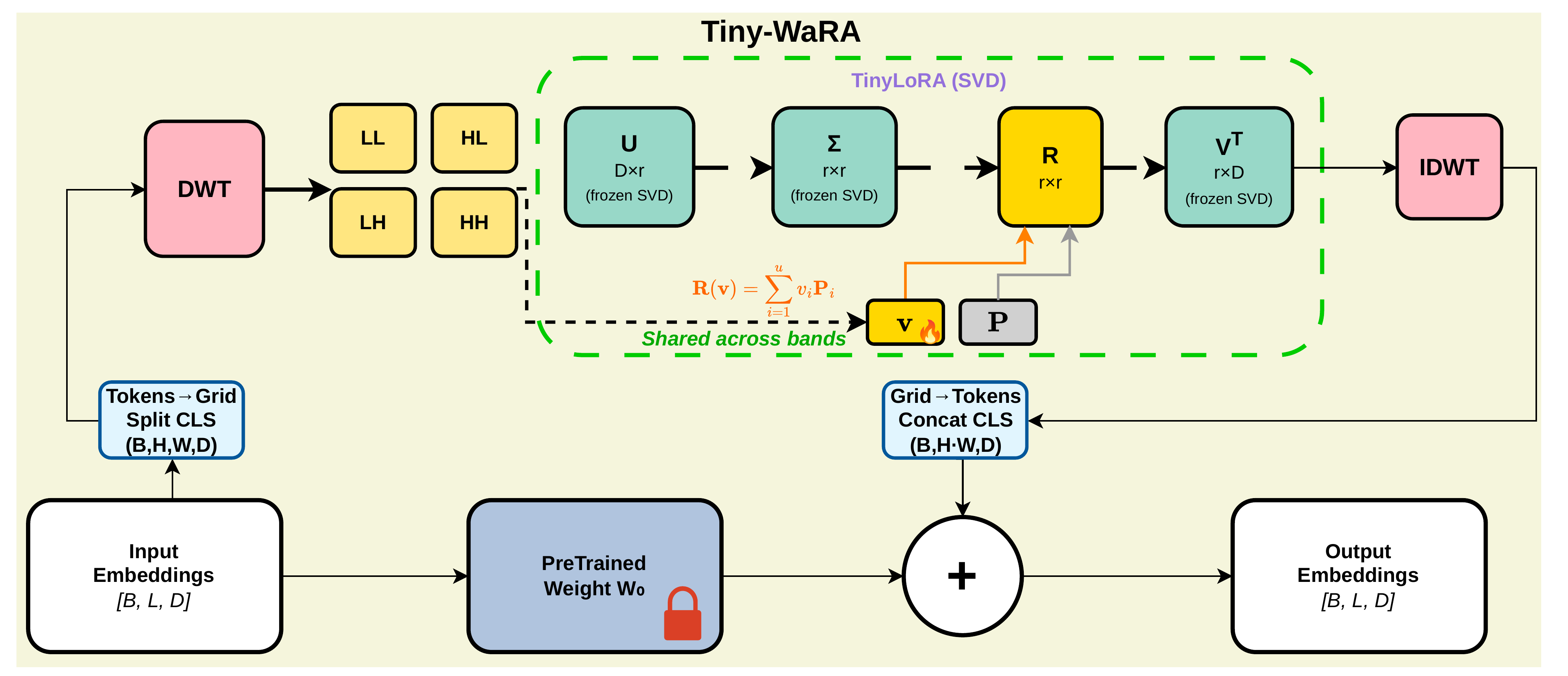}
  \caption{Tiny-WaRA.}
  \label{fig:overview_tiny}
\end{subfigure}
\caption{\textbf{Overview of WaRA and Tiny-WaRA.} (a) WaRA performs low-rank
adaptation in a wavelet domain by applying DWT on patch tokens, updating subband
coefficients with shared adapter parameters, and reconstructing an additive
update via IDWT. (b) Tiny-WaRA replaces the standard adapter with a compact
parameterization that learns only a small coefficient vector in a fixed basis
derived from pretrained weights~\cite{balazy2024lora,morris2026learning}.}
\label{fig:Model}
\end{figure}

The core idea of WaRA, shown in \cref{fig:Model}, is to perform parameter-efficient adaptation in a wavelet domain that exposes multi-scale structure while keeping the pretrained layer frozen. Consider a linear projection with frozen pretrained weights $W_{0}\in\mathbb{R}^{d_{2}\times d_{1}}$ applied to token embeddings $X\in\mathbb{R}^{B\times L\times d_{1}}$ ($B$: Batch Size, $L$: number of patch token + class token , $d_1$: Embedding size) which produces the base output $Y_{\text{base}} = XW_{0}^{\top}$. LoRA adds an additive update using low-rank factors, that is $\Delta W = BA$ with $A\in\mathbb{R}^{r\times d_{1}}$ and $B\in\mathbb{R}^{d_{2}\times r}$, where $r \ll d_1, d_2$ \cite{hu2022lora}. 
Inspired by LoRA, WaRA applies low-rank adaptation in the multi-scale wavelet domain rather than in the feature domain. Specifically, we split $X$ into the class token $X_{\text{cls}}$ and patch tokens $X_{\text{tok}}$, reshape the patch tokens into a spatial grid $X_{\text{tok}_r} \in \mathbb{R}^{B \times H \times W \times d_1}$, and then apply a fixed 2D wavelet transform:

\begin{equation}
X = [X_{\text{cls}}, X_{\text{tok}}],\qquad
C=\mathcal{W}(X_{\text{tok}_r})=\{C_{\text{LL}},C_{\text{LH}},C_{\text{HL}},C_{\text{HH}}\},
\end{equation}
where $C_{\alpha}\in\mathbb{R}^{B\times H'\times W'\times d_{1}}$ and $(H',W')=(H/2,W/2)$ for one level of decomposition, with $\alpha$ indexing the wavelet subbands (e.g., $\mathrm{LL}$, $\mathrm{LH}$, $\mathrm{HL}$, $\mathrm{HH}$).
 In the wavelet domain, we apply a shared low-rank adapter along the feature dimension for every subband:
\begin{equation}
\Delta C_{\alpha} = C_{\alpha}A^{\top}B^{\top},\qquad
A\in\mathbb{R}^{r\times d_{1}},\quad B\in\mathbb{R}^{d_{2}\times r}.
\end{equation}
Following the standard LoRA initialization \cite{hu2022lora}, we initialize $A$ with a Gaussian distribution and $B$ with zeros. Parameter sharing across the four subbands avoids the parameter growth that would arise from using independent adapters per subband. We reconstruct the patch token update via the inverse transform, and we adapt the class token with the same adapter:
\begin{equation}
\Delta Y_{\text{tok}}=\mathcal{W}^{-1}(\{\Delta C_{\alpha}\}),\qquad
\Delta Y_{\text{cls}} = X_{\text{cls}}A^{\top}B^{\top}.
\end{equation}
Finally, the layer output is formed by residual fusion,
$Y = Y_{\mathrm{base}} + [\Delta Y_{\mathrm{cls}}, \Delta Y_{\mathrm{tok}}]$.

\subsection{Tiny-WaRA for extreme low resource adaptation}
LoRA is efficient, but storing and training adapters can still be costly when many specializations are required or when the adaptation budget is extremely small. Tiny-WaRA reduces the trainable budget even more by parameterizing the adapter update with a compact trainable core on a fixed basis derived from $W_{0}$ \cite{balazy2024lora,morris2026learning}. For each layer, we compute Singular Value Decomposition (SVD) and then truncate it to rank r by only keeping the r-largest singular values of pretrained weights:
\begin{equation}
W_{0} \approx U\,\mathrm{diag}(S)\,V^{\top},\qquad
U\in\mathbb{R}^{d_{2}\times r},\; S\in\mathbb{R}^{r},\; V\in\mathbb{R}^{d_{1}\times r}.
\end{equation}
We then define a mixing matrix as a linear combination of fixed basis matrices $\{P_i\}_{i=1}^{u}$ (initialized with a Gaussian distribution) with a trainable coefficient vector $v\in\mathbb{R}^{u}$ (with a zero initialization):
\begin{equation}
R(v)=\sum_{i=1}^{u} v_i P_i,\qquad
\Delta W(v)=U\,\mathrm{diag}(S)\,R(v)\,V^{\top},
\end{equation}
so only $v$ is trainable. Tiny-WaRA uses the same wavelet pipeline as WaRA, but replaces the subband update with $\Delta W(v)$:
\begin{equation}
\Delta C_{\alpha} = C_{\alpha}\Delta W(v)^{\top},\qquad
\Delta Y_{\text{cls}} = X_{\text{cls}}\Delta W(v)^{\top},
\end{equation}
followed by IDWT reconstruction and residual fusion as above.

\section{Results}

\textbf{Datasets, Training and Evaluation Setup.}
We evaluate on four medical image classification datasets, namely Messidor-2 \cite{decenciere2014feedback}, ISIC2018 \cite{codella2019skin}, DDSM \cite{lee2017curated}, and CXR COVID \cite{siddhartha2021covid}, with 1,740, 11,527, 9,104, and 1,708 images, respectively. These correspond to 5-class diabetic retinopathy grading, 7-class skin lesion classification, 3-class mammography classification, and 3-class chest radiograph classification, respectively. For each dataset, we use the provided split directory with separate train, val, and test folders. We initialize the vision backbone with the BiomedCLIP ViT-B model using $16 \times 16$ patches, and replace the classification head to match the dataset's label space. All images are resized to $224 \times 224$ and trained for 20 epochs with a batch size of 16. We use AdamW with weight decay 0.1 and a cosine learning rate schedule. We use a learning rate of $2\times10^{-5}$ for full fine tuning, $2\times10^{-4}$ for attention tuning, and $2\times10^{-3}$ for the remaining parameter efficient methods, including WaRA. We select the best checkpoint using validation AUC and train all models on a 16 GB NVIDIA V100  GPU. For all baselines, we follow the training configurations from their official repositories, and for FourierFT \cite{gao2024parameter}, LoCA \cite{duloca}, and FouRA \cite{borse2024foura}, for matched parameter budget comparisons in Tiny-WaRA experiments (\cref{tab:tiny_wara}), we shrink their adaptation capacity to match the Tiny-WaRA parameter budget. Specifically, we reduce FourierFT by decreasing the number of learned Fourier components from 600 to 360, LoCA by lowering its low-rank adaptation size from 600 to 262, and FouRA by setting its rank to 64. For Tiny-WaRA, we use rank $r=64$ with basis size $u=128$.

\begin{table}[t]
\centering
\footnotesize
\setlength{\tabcolsep}{2.0pt}
\renewcommand{\arraystretch}{1.10}
\caption{Comparison results across four evaluation datasets. Per dataset columns report \textbf{F1} (\%). `Params.' denotes the ratio of learnable parameters to the total number of parameters. Entries are reported as mean$\pm$std. Among non-gray rows, the best and second-best means are colored blue and red, respectively.}
\label{tab:biomedclip_f1_std}

\begin{adjustbox}{max width=0.9\linewidth,center}
\begin{tabular}{@{}L{0.30\linewidth}c c c c c c@{}}
\toprule
Method & Params. &
\makecell[c]{Fundus\\(Messidor-2)} &
\makecell[c]{Dermoscopy\\(ISIC2018)} &
\makecell[c]{Mammography\\(DDSM)} &
\makecell[c]{Chest X-ray\\(COVID)} &
\makecell[c]{Avg.\\F1} \\
\midrule
\textcolor{gray}{Full fine tuning} & \textcolor{gray}{100} &
\textcolor{gray}{39.04$\pm$5.01} & \textcolor{gray}{67.46$\pm$1.17} & \textcolor{gray}{72.19$\pm$0.71} & \textcolor{gray}{97.50$\pm$0.40} &
\textcolor{gray}{69.05$\pm$1.82} \\
\textcolor{gray}{Linear probing} & \textcolor{gray}{0.0045} &
\textcolor{gray}{28.40$\pm$3.90} & \textcolor{gray}{62.49$\pm$1.17} & \textcolor{gray}{61.54$\pm$1.70} & \textcolor{gray}{97.56$\pm$0.20} &
\textcolor{gray}{62.50$\pm$1.74} \\
\textcolor{gray}{Bias} & \textcolor{gray}{0.12} &
\textcolor{gray}{34.84$\pm$4.39} & \textcolor{gray}{64.78$\pm$1.49} & \textcolor{gray}{65.48$\pm$0.90} & \textcolor{gray}{97.83$\pm$0.23} &
\textcolor{gray}{65.73$\pm$1.75} \\
\midrule
Prompt tuning \cite{jia2022visual} & 0.176 &
30.58$\pm$3.88 & 62.94$\pm$1.02 & 62.38$\pm$2.35 & 97.36$\pm$0.26 &
63.31$\pm$1.88 \\
Attention tuning \cite{touvron2022three} & 33.04 &
30.42$\pm$8.18 & 63.26$\pm$2.23 & 62.88$\pm$1.21 & 97.31$\pm$0.56 &
63.47$\pm$3.05 \\
Adapter \cite{houlsby2019parameter} & 2.05 &
25.68$\pm$9.67 & 67.22$\pm$2.12 & 65.97$\pm$2.61 & \textcolor{blue}{\textbf{97.62$\pm$0.38}} &
64.12$\pm$3.69 \\
LoRA (rank=16) \cite{hu2022lora} & 0.69 &
39.69$\pm$3.07 & \textcolor{blue}{\textbf{71.81$\pm$1.10}} & 68.22$\pm$7.04 & 96.40$\pm$0.84 &
69.03$\pm$3.01 \\
LoRA (rank=8) & 0.34 &
\textcolor{red}{40.50$\pm$4.55} & 70.61$\pm$0.92 & \textcolor{red}{73.96$\pm$1.77} & 96.96$\pm$0.38 &
\textcolor{red}{70.51$\pm$1.90} \\
FourierFT \cite{gao2024parameter} & 0.16 &
18.63$\pm$1.03 & 39.90$\pm$1.18 & 62.24$\pm$1.91 & 92.44$\pm$0.95 &
53.30$\pm$1.27 \\
LoCA \cite{duloca} & 0.17 &
31.36$\pm$2.44 & 58.11$\pm$2.02 & 62.65$\pm$1.22 & 95.93$\pm$0.38 &
62.01$\pm$1.52 \\
FouRA \cite{borse2024foura} & 0.68 &
37.60$\pm$4.63 & 71.30$\pm$1.22 & 67.47$\pm$4.20 & \textcolor{red}{97.58$\pm$0.42} &
68.49$\pm$1.66 \\
WaveFT \cite{bilican2025exploring} & 0.17 &
17.99$\pm$1.42 & 32.93$\pm$3.16 & 44.47$\pm$4.81 & 90.32$\pm$0.64 &
46.43$\pm$2.51 \\
\midrule
WaRA (rank=8) & 0.34 &
\textcolor{blue}{\textbf{45.35$\pm$2.60}} & 70.94$\pm$0.99 & \textcolor{blue}{\textbf{74.68$\pm$0.94}} & 96.81$\pm$0.55 &
\textcolor{blue}{\textbf{71.94$\pm$1.27}} \\
WaRA (rank=16) & 0.69 &
38.73$\pm$7.07 & \textcolor{red}{71.44$\pm$1.05} & 70.20$\pm$2.61 & 96.64$\pm$0.26 &
69.25$\pm$2.75 \\
\bottomrule
\end{tabular}
\end{adjustbox}
\end{table}
\begin{table}[t]
\centering
\footnotesize
\setlength{\tabcolsep}{1.7pt}
\renewcommand{\arraystretch}{1.10}
\caption{Average performance across the four datasets (\%). Entries are reported as mean$\pm$std, where the std term is the average of per-dataset stds. Among non-gray rows, the best and second-best means are colored blue and red, respectively.}
\label{tab:biomedclip_avg_std}
\begin{adjustbox}{max width=0.9\linewidth,center}
\begin{tabular}{@{}L{0.30\linewidth}c c c c c c c@{}}
\toprule
Method & Params. & F1 & ACC & AUC & Recall & Precision & Kappa \\
\midrule
\textcolor{gray}{Full fine tuning} & \textcolor{gray}{100} &
\textcolor{gray}{69.05$\pm$1.82} & \textcolor{gray}{78.12$\pm$1.07} & \textcolor{gray}{90.50$\pm$0.42} &
\textcolor{gray}{67.75$\pm$1.80} & \textcolor{gray}{74.12$\pm$4.00} & \textcolor{gray}{73.43$\pm$1.15} \\
\textcolor{gray}{Linear probing} & \textcolor{gray}{0.0045} &
\textcolor{gray}{62.50$\pm$1.74} & \textcolor{gray}{72.61$\pm$2.72} & \textcolor{gray}{86.71$\pm$0.35} &
\textcolor{gray}{62.54$\pm$2.02} & \textcolor{gray}{66.77$\pm$2.24} & \textcolor{gray}{64.89$\pm$2.89} \\
\textcolor{gray}{Bias} & \textcolor{gray}{0.12} &
\textcolor{gray}{65.73$\pm$1.75} & \textcolor{gray}{75.78$\pm$0.86} & \textcolor{gray}{88.76$\pm$0.36} &
\textcolor{gray}{64.99$\pm$2.05} & \textcolor{gray}{68.99$\pm$2.33} & \textcolor{gray}{69.50$\pm$2.63} \\
\midrule
Prompt tuning \cite{jia2022visual} & 0.176 &
63.31$\pm$1.88 & 74.34$\pm$1.00 & 86.90$\pm$0.28 &
62.45$\pm$1.83 & 67.18$\pm$2.36 & 66.68$\pm$1.67 \\
Attention tuning \cite{touvron2022three} & 33.04 &
63.47$\pm$3.05 & 73.54$\pm$1.28 & 86.76$\pm$1.08 &
63.23$\pm$2.76 & 67.02$\pm$4.89 & 66.13$\pm$4.02 \\
Adapter \cite{houlsby2019parameter} & 2.05 &
64.12$\pm$3.69 & 75.47$\pm$1.74 & 87.38$\pm$2.10 &
63.57$\pm$3.62 & 67.27$\pm$4.58 & 64.35$\pm$6.68 \\
LoRA (rank=16) \cite{hu2022lora} & 0.69 &
69.03$\pm$3.01 & 76.91$\pm$2.97 & 90.32$\pm$1.14 &
68.27$\pm$2.87 & 73.72$\pm$3.15 & 73.03$\pm$3.69 \\
LoRA (rank=8) & 0.34 &
\textcolor{red}{70.51$\pm$1.90} & \textcolor{red}{78.71$\pm$1.20} & \textcolor{red}{91.14$\pm$0.46} &
\textcolor{red}{70.17$\pm$2.66} & \textcolor{red}{74.52$\pm$2.42} & \textcolor{red}{75.62$\pm$2.47} \\
FourierFT \cite{gao2024parameter} & 0.16 &
53.30$\pm$1.27 & 71.69$\pm$0.87 & 84.15$\pm$1.49 &
53.08$\pm$1.00 & 59.72$\pm$3.36 & 53.98$\pm$2.29 \\
LoCA \cite{duloca} & 0.17 &
62.01$\pm$1.52 & 74.26$\pm$0.75 & 87.49$\pm$0.26 &
60.98$\pm$1.47 & 67.30$\pm$1.73 & 64.43$\pm$2.10 \\
FouRA \cite{borse2024foura} & 0.68 &
68.49$\pm$1.66 & 77.28$\pm$0.98 & 89.73$\pm$0.85 &
67.57$\pm$1.64 & 72.74$\pm$1.76 & 72.39$\pm$2.12 \\
WaveFT \cite{bilican2025exploring} & 0.17 &
46.43$\pm$2.51 & 66.36$\pm$2.59 & 80.42$\pm$1.04 &
48.04$\pm$1.40 & 51.25$\pm$3.50 & 44.30$\pm$4.75 \\
\midrule
WaRA (rank=8) & 0.34 &
\textcolor{blue}{\textbf{71.94$\pm$1.27}} & \textcolor{blue}{\textbf{79.69$\pm$0.90}} & \textcolor{blue}{\textbf{91.62$\pm$0.37}} &
\textcolor{blue}{\textbf{71.59$\pm$1.78}} & \textcolor{blue}{\textbf{75.02$\pm$2.27}} & \textcolor{blue}{\textbf{77.01$\pm$1.48}} \\
WaRA (rank=16) & 0.69 &
69.25$\pm$2.75 & 77.22$\pm$1.60 & 90.43$\pm$0.79 &
69.05$\pm$2.60 & 72.83$\pm$3.58 & 72.81$\pm$3.15 \\
\bottomrule
\end{tabular}
\end{adjustbox}
\end{table}
\subsection{Comparisons with State-of-the-art}
\Cref{tab:biomedclip_f1_std,tab:biomedclip_avg_std} compare WaRA against representative parameter-efficient transfer learning baselines on four medical image classification datasets. Full fine-tuning provides a strong reference but updates all parameters, whereas linear probing is a low-cost lower bound and consistently underperforms. We report F1 scores in \Cref{tab:biomedclip_f1_std}, where higher values indicate a better balance between sensitivity and precision at the operating threshold, a clinically relevant criterion that reflects improved decision-level performance and fewer false alarms.

\noindent According to \cref{tab:biomedclip_f1_std}, WaRA (rank=8) achieves the best average performance among all methods, outperforming all parameter-efficient baselines. Compared to LoRA (rank=8), our strongest competitor, WaRA, improves the average F1 by +1.43 (71.94 vs.\ 70.51). The gains are most pronounced on Fundus (+4.85) and remain positive on Mammography (+0.72), while WaRA is slightly lower on Dermoscopy (-0.33) and comparable on Chest X-ray with a small difference (-0.81). We further report both ranks 8 and 16 for LoRA and WaRA; in both cases, rank=8 performs better overall, indicating that increasing the rank does not necessarily improve performance.\\
\noindent Additionally, \cref{tab:biomedclip_avg_std} reports performance averaged across the four datasets over multiple metrics: F1, accuracy (ACC), area under the ROC curve (AUC), Recall, Precision, and Cohen's $\kappa$, which corrects for chance agreement
and is therefore more informative than accuracy under class imbalance.
Consistent with \cref{tab:biomedclip_f1_std}, WaRA (rank=8) outperforms LoRA (rank=8) across all metrics in \cref{tab:biomedclip_avg_std}, while also exhibiting lower standard deviation, indicating more stable performance across runs.\\
\noindent We also compare against recent frequency-domain adaptation baselines, including FourierFT \cite{gao2024parameter}, LoCA \cite{duloca}, FouRA \cite{borse2024foura}, and the wavelet-domain method WaveFT \cite{bilican2025exploring}, which was originally proposed for personalized text-to-image generation. FouRA is the strongest among these methods; nevertheless, WaRA (rank=8) improves over FouRA by substantial margins (e.g., +3.45 F1, +2.41 ACC, +1.89 AUC, +4.02 Recall, +2.28 Precision, and +4.62 $\kappa$). Overall, these results suggest that wavelet-based adaptation outperforms both feature-space and Fourier-space parameterizations, which we attribute to multi-resolution parameter updates that capture both global structure and localized details more effectively than Fourier-based updates.\\
\noindent We next consider an ultra-low parameter regime in Table~\ref{tab:tiny_wara}, where all methods are matched to a comparable trainable parameter budget by reducing their adaptation capacity. Tiny-WaRA trains a compact coefficient vector of size $u=128$ over a fixed basis while using rank $r=64$, which yields only $0.0078\%$ trainable parameters. Despite this severe constraint, Tiny-WaRA achieves an average AUC of $88.81$ with $78.54$ ACC and $69.43$ F1, and it remains the top performer across the four datasets. This setting is practically important for medical image classification as labeled data can be limited, and adaptation is often performed under tight memory, compute, and communication budgets, including on-premises deployment or privacy-preserving federated training, where the update size directly affects bandwidth and system cost \cite{dutt2023parameter}.
\begin{table}[t]
    \centering
    \renewcommand{\arraystretch}{1.2}
    \caption{Comparison results of Tiny-WaRA with selected PEFT baselines across four evaluation datasets. Per dataset columns report \textbf{AUC} (\%). `Params.' denotes the ratio of learnable parameters to the total number of parameters. The best and second best results are bold and underlined, respectively.}
    \resizebox{\textwidth}{!}{%
    \begin{tabular}{lcccccccc}
    \toprule
    \multirow{2}{*}{Method} & \multicolumn{1}{c}{Computing cost} &
    \multirow{2}{*}{\begin{tabular}[c]{@{}c@{}}Fundus\\ (Messidor-2)\end{tabular}} &
    \multirow{2}{*}{\begin{tabular}[c]{@{}c@{}}Dermoscopy\\ (ISIC2018)\end{tabular}} &
    \multirow{2}{*}{\begin{tabular}[c]{@{}c@{}}Mammography\\ (DDSM)\end{tabular}} &
    \multirow{2}{*}{\begin{tabular}[c]{@{}c@{}}Chest X-ray\\ (COVID)\end{tabular}} &
    \multicolumn{3}{c}{Performance (Avg.)} \\ \cline{7-9}
    & Params. & & & & & AUC & ACC & F1 \\
    \midrule
    FourierFT \cite{gao2024parameter} & 0.010 & 53.18 & 61.97 & 60.36 & 63.17 & 59.67 & 46.24 & 24.90 \\
    LoCA \cite{duloca} & 0.010  & 79.65 & 92.25 & 79.16 & 99.04 & 87.53 & 74.22 & 62.30 \\
    FouRA \cite{borse2024foura} & 0.008 & 78.82 & 93.41 & 80.73 & 99.01 & 87.99 & 75.11 & 65.04 \\
    \midrule
    Tiny-WaRA (Ours) & 0.0078 & \textbf{81.05} & \textbf{93.48} & \textbf{81.47} & \textbf{99.22} & \textbf{88.81} & \textbf{78.54} & \textbf{69.43} \\
    \bottomrule
    \end{tabular}%
    }
    \label{tab:tiny_wara}
\end{table}

\subsection{Analysis and Ablation Study}
\begin{figure*}[t]
\centering

\begin{subfigure}[t]{0.45\textwidth}
  \centering
  \includegraphics[width=\linewidth,trim=6 6 6 6,clip]{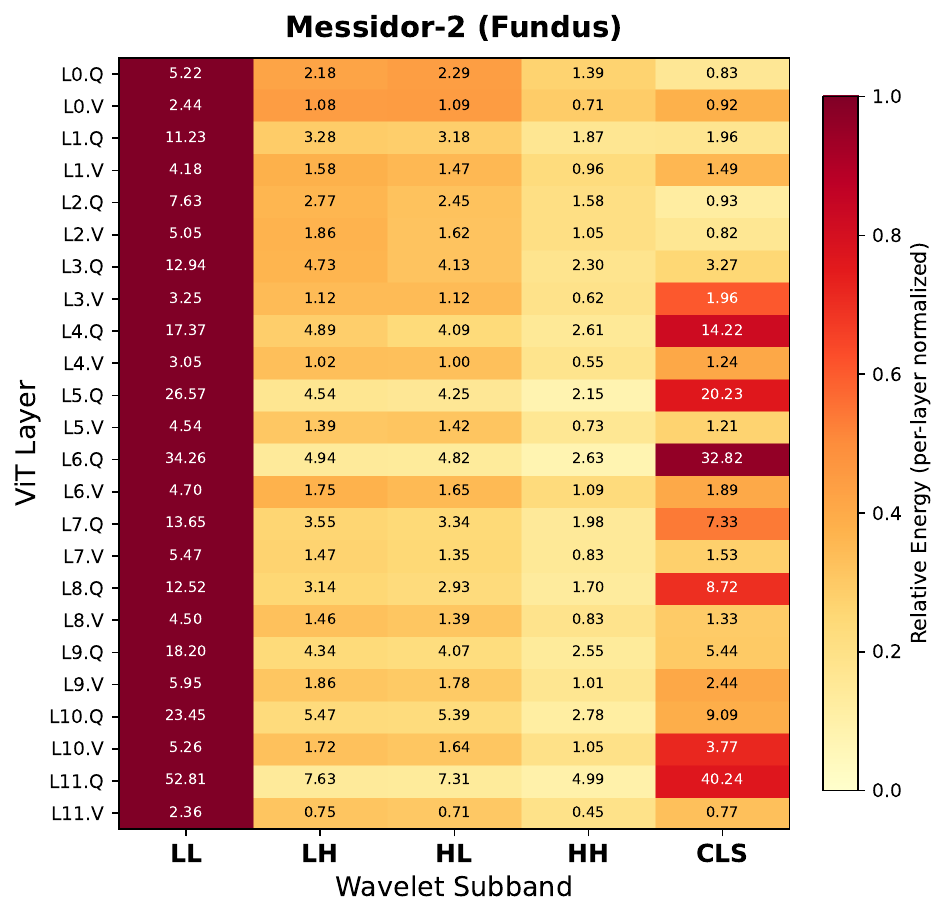}
  \caption{}
  \label{fig:subband}
\end{subfigure}
\hfill
\begin{subfigure}[t]{0.51\textwidth}
  \centering
  \includegraphics[width=\linewidth]{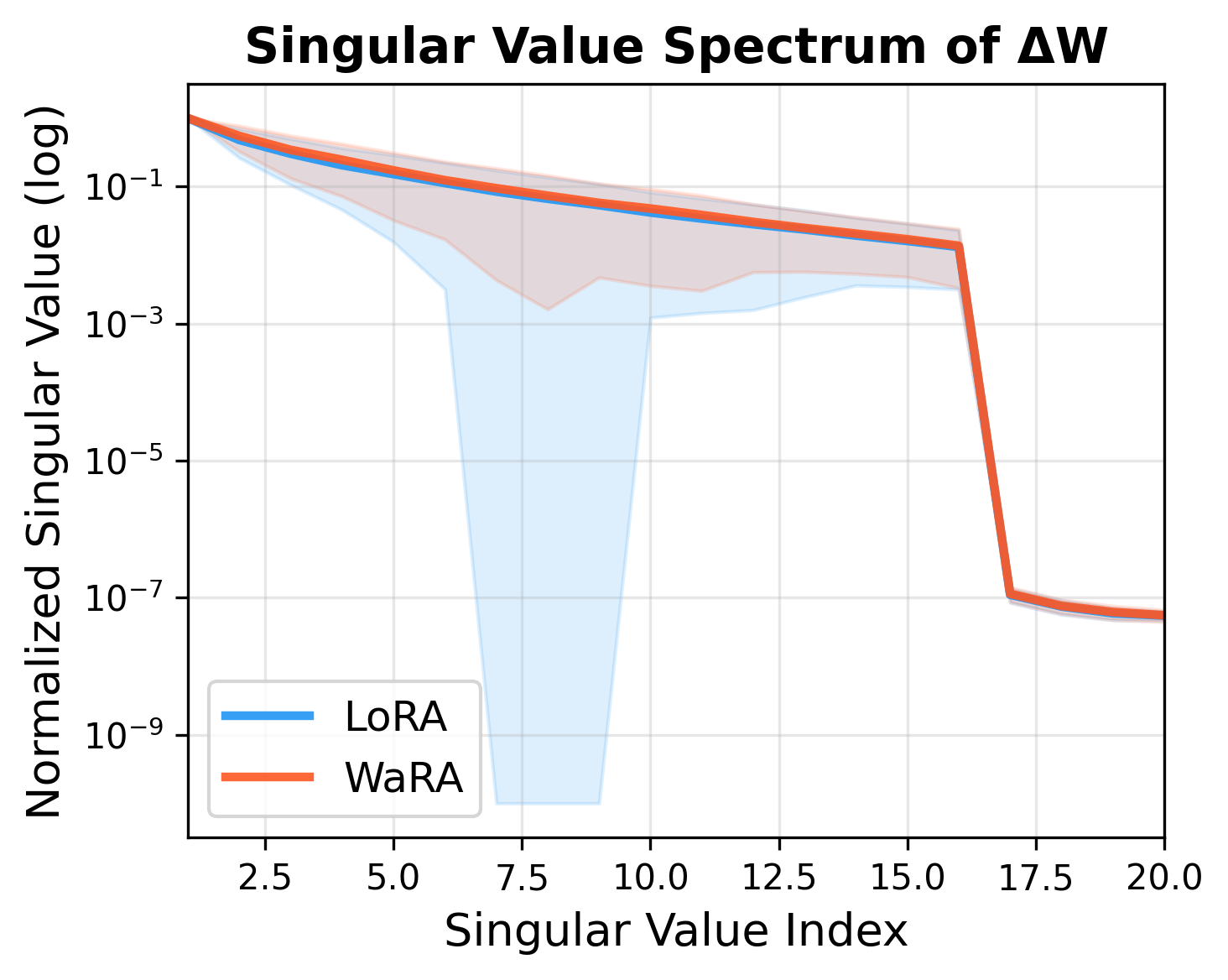}
  \caption{}
  \label{fig:svd_spect}
\end{subfigure}

\caption{Complementary analysis of WaRA. (a): Wavelet subband energy pattern on Fundus.  (b): Singular value spectrum of the corresponding weight update.}
\label{fig:wavelet_radar_svd}
\end{figure*}

\textbf{Wavelet energy across subbands.}
\Cref{fig:subband} reports the relative wavelet subband energy of the learned update across ViT layers on the Fundus dataset. As shown, the LL band consistently carries the majority of the update energy, while the high frequency bands (LH, HL, HH) remain smaller but non-zero. This supports our design choice to fix and share the update parameterization across subbands, since allocating separate large capacity per band would be inefficient given the observed energy imbalance. At the same time, the presence of structured energy in the higher frequency bands, together with nontrivial CLS energy, indicates that the model still benefits from a fused multi-resolution update, where coarse structure is primarily adjusted through LL, and finer cues are injected through the remaining bands via the inverse wavelet reconstruction.
\\
\textbf{Singular value spectrum of the update.}
\Cref{fig:svd_spect} compares the singular value spectra of the effective weight
updates ($\Delta W$) of LoRA and WaRA across all evaluated layers and datasets
(solid lines: mean; shaded areas: cross-layer and cross-dataset variance). LoRA's
singular values decay rapidly, concentrating update capacity in a small subset of
dominant directions, consistent with existing literature describing LoRA updates as highly anisotropic \cite{song2024sharelora}. WaRA, by contrast, yields a notably flatter spectrum, indicating that its update energy is distributed more evenly across directions, which is consistent with the lower run-to-run variance observed in \Cref{tab:biomedclip_avg_std}.
\\
\textbf{Shared vs. band-specific adapters across wavelet subbands.}
In WaRA, we apply the same low-rank adapter parameters across all four wavelet subbands. A natural alternative is to assign independent LoRA adapters for different subbands, i.e., using separate $(A_\alpha, B_\alpha)$ for each $\alpha \in \{\text{LL}, \text{LH}, \text{HL}, \text{HH}\}$, instead of sharing a single $(A,B)$ across bands. Table~\ref{tab:ablation_shared_vs_multi} reports the ablation results. Sharing one adapter across subbands yields consistently better performance on Messidor-2, ISIC2018, and DDSM, while achieving comparable performance on COVID. Overall, the shared design improves the average AUC from 90.77 to 91.62 (+0.85), while also being more resource efficient due to the substantially reduced trainable footprint and optimizer overhead. These results support our design choice of subband sharing, indicating that the wavelet decomposition already provides complementary multi-resolution pathways, and allocating separate adapter capacity per band is unnecessary under the same training setup. 

\begin{table}[h]
\centering
\caption{Ablation study on wavelet-subband adapter design (AUC, \%).}
\setlength{\tabcolsep}{10pt}
\resizebox{\textwidth}{!}{%
\begin{tabular}{lcccc}
\toprule
Method & Messidor-2 & ISIC2018 & DDSM & COVID \\
\midrule
WaRA (shared adapter across bands) & 83.85 & 95.29 & 87.73 & 99.61 \\
WaRA (band-specific adapters)      & 83.35 & 94.21 & 85.84 & 99.68 \\
\bottomrule
\end{tabular}
}
\label{tab:ablation_shared_vs_multi}
\end{table}

\section{Impact in Resource-Constrained Settings}
\label{sec:rcs}

Deploying foundation models in resource-constrained settings is limited less by
attainable accuracy than by data scarcity and limited computational resources.
WaRA addresses both: adapting only $0.34\%$ of parameters, it outperforms every
fine-tuning strategy, including full fine-tuning and Fourier-domain
adaptation, and our Tiny-WaRA variant retains the best performance at a matched
ultra-low budget of $0.0078\%$. A WaRA adapter occupies $1.2$\,MB and a Tiny-WaRA adapter $12$\,KB, against
$330$\,MB for full fine-tuning. A single frozen backbone can therefore be
deployed once and specialized to new tasks or sites by exchanging adapters small
enough for intermittent or metered connections. The same reduction lowers the
per-round uplink in federated training by two orders of magnitude relative to
LoRA, and as only a low-dimensional update leaves the institution, the
scheme remains applicable where raw data cannot be shared \cite{dutt2023parameter}.

\section{Conclusion}
\label{sec:conclusion}

We introduced WaRA, a wavelet-domain low-rank adaptation module for medical image
classification, together with Tiny-WaRA for extremely low-parameter regimes. Across
four modalities, WaRA improves over strong parameter-efficient baselines at a
matched budget, and Tiny-WaRA outperforms matched-budget frequency-domain methods while training only $0.0078\%$ of parameters, making it well-suited to deployment under tight storage, compute, and bandwidth constraints. Future work includes extending WaRA to multiple backbones and dense prediction tasks such as segmentation and detection, and exploring multi-level decompositions and alternative wavelet families.

\subsubsection{Acknowledgments.}
This work was supported by the Canadian Foundation for Innovation-John R. Evans Leaders Fund (CFI-JELF) program grant number 42816. Mitacs Accelerate program grant number AWD024298-IT33280. We also acknowledge the support of the Natural Sciences and Engineering Research Council of Canada (NSERC), [RGPIN-2023-03575]. Cette recherche a été financée par le Conseil de recherches en sciences naturelles et en génie du Canada (CRSNG), [RGPIN-2023-03575].

\subsubsection{Disclosure of Interests.} The authors have no competing interests in the paper.

\bibliographystyle{splncs04}
\bibliography{ref}

\end{document}